# The empirical structure of word frequency distributions

Michael Ramscar

*Eberhard Karls Universität Tübingen*

**The frequencies at which individual words occur across languages follow power law distributions, a pattern of findings known as Zipf's law.[1,2,3] A vast literature argues over whether this serves to optimize the efficiency of human communication,[4,5,6,7] however this claim is necessarily post hoc,[8] and it has been suggested that Zipf's law may in fact describe mixtures of other distributions.[9,10] From this perspective, recent findings that Sinosphere first (family) names are geometrically distributed[11,12,13,14] are notable, because this is actually consistent with information theoretic predictions regarding optimal coding.[15,16] First names form natural communicative distributions in most languages, and I show that when analyzed in relation to the communities in which they are used, first name distributions across a diverse set of languages are both geometric and, historically, remarkably similar, with power law distributions only emerging when empirical distributions are aggregated. I then show this pattern of findings replicates in communicative distributions of English nouns and verbs. These results indicate that if lexical distributions support efficient communication, they do so because their functional structures directly satisfy the constraints described by information theory, and not because of Zipf's law. Understanding the function of these information structures is likely to be key to explaining humankind's remarkable communicative capacities.**

Variable-length coding allows lossless source codes to be compressed arbitrarily close to their entropy, enabling communication efficiency to be maximized close to its theoretical limit.[17] [18] The benefits of these codes can be illustrated as follows: if *one* is easier to articulate than *four*, and *twenty* is harder than *four*, it follows that using *one* more often than *four*, and *twenty* less, will reduce the articulatory cost of communicating numbers. Empirically, shorter, less informative words (*one*) are indeed more frequent than longer, more informative words (*twenty*),[1,2,3,19] such that across languages, type / token frequency distributions follow Zipf's law: a long-tailed power law distribution in which around half of the tokens represent a small number of shorter, high-frequency types ('*and*', '*the*'), with the other half representing very large numbers of longer, low-frequency types ('*comprise*', '*corpus*'). It is thus often claimed that languages are, in some sense, variable length codes, and that the power law distributions of words serve to optimize communication. [3,4,5,6,7]

There are, however, reasons to doubt these claims. First, linguistic fits to power laws are often poor, and better described by other distributions.[20] Second, power law distributions can simply represent mixtures of other distributions.[9] [10] Third, results in coding and information theory show that, in fact, members of the exponential family of distributions maximize the communicative efficiency of codes.[15] [16] Fourth, at least some lexical distributions, such as first names in the unrelated languages of Korean and Chinese (in both languages, family names come first in speech) violate Zipf's law, because they have been shown to be exponentially distributed [11] [12] [13] [14] (Figure 1A; because names are discrete, strictly speaking these are geometric distributions).

Vietnamese is unrelated to Korean and Chinese, but its family names also occur first in speech. Since most Vietnamese-Americans sampled by the 2000 US census were named in Vietnam, a partial Vietnamese family name distribution could be recreated to explore the generalizability of the Chinese and Korean findings. This distribution is highly similar to Korean, and also approximates a geometric (Figure 1, see supplemental materials).

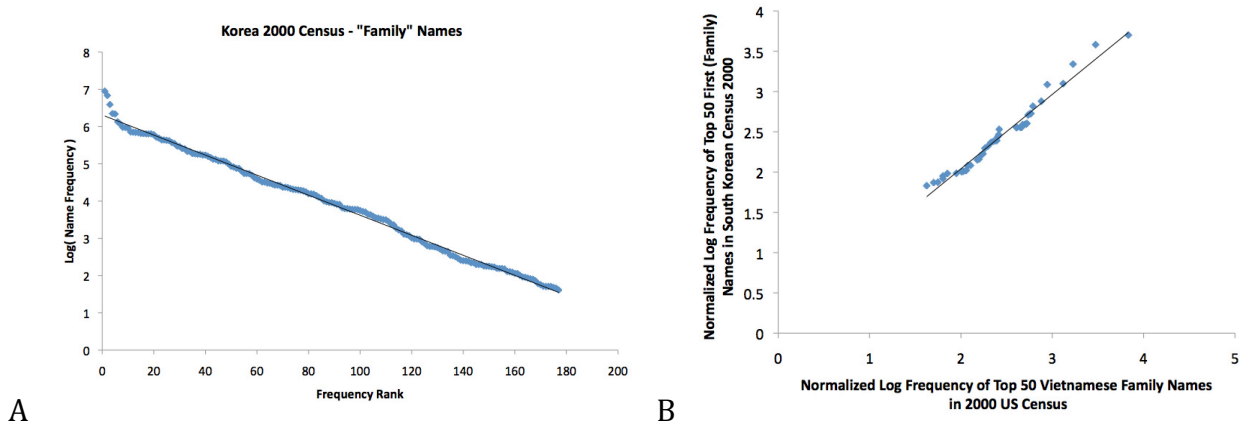

**Figure 1A** Log frequency x frequency rank plots (linear = exponential) of family names (minimum frequency = 40) in South Korean Census 2000 ($R^2$=.99). **1 B.** Frequency normalized comparison of the Korean distribution with a sample Vietnamese family name taken from the 2000 US Census ($R^2$=.97).

Historically, most people tended to have a single name, with bynames being used to create more specific identifiers.[21] [22] Sinosphere family names derive from regulations that transformed male first names into patronyms,[23] whereas bynames were turned into patronyms in the West.[22] In speech, full Western and Sinosphere names are sequences of tokens, in which Western given names and Sinosphere family names come first. Both of these considerations suggest that the appropriate communicative comparators to Sinosphere family names are Western given names, as do analyses of historical name distributions in England, where in every 50-year period 1550 - 1799 50% of girls shared 3 feminine names, and 50% of boys 3 masculine names.[24] Sinosphere first name distributions changed little in the 19th Century,[13] whereas after UK naming was standardized in the Births and Deaths Registration Act (1836), its first name distribution changed considerably. Figure 2 plots the decline in the popularity of the most frequent 3 male ($r^2$=-.98) and female ($r^2$=-.997) names in this period, suggesting that naming was highly sensitive to informativity (more identities require more names) in this system.

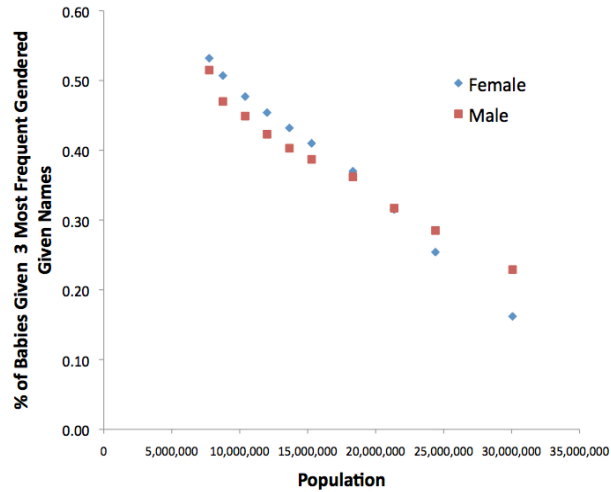

**Figure 2** The frequency with which the top three male and female names were given in England[25] by decade 1800 – 1900 plotted against the by-decade UK population[26][27] 1801-1901 (1890 data unavailable).

To estimate the distribution of UK first names prior to these changes, data from 23,935 name registrations made in 4 parishes in Scotland between 1701–1800 were analyzed,[28] revealing that first names had a geometric distribution (Figure 3A, 3B and 3C) both by gender and when aggregated across genders. These results were replicated on a set of birth records (25,993 names) for the English counties of Durham and Northumberland over the same period [29] (Figure 3D; see supplemental materials). Given that historical name stocks in both China and Korea comprised 100 names [13][23] (the colloquial Chinese expression for the common people – '*Bǎijiāxìng*' – means 'the hundred names'), it is notable that the name stock in each Scottish parish was also around 100, such that both the distribution ($r^2$=.96) and information entropy (Korea=4.7 bits; Scotland=4.8 bits) of historical Scottish and modern Korean first names are remarkably similar (Figure 3E).

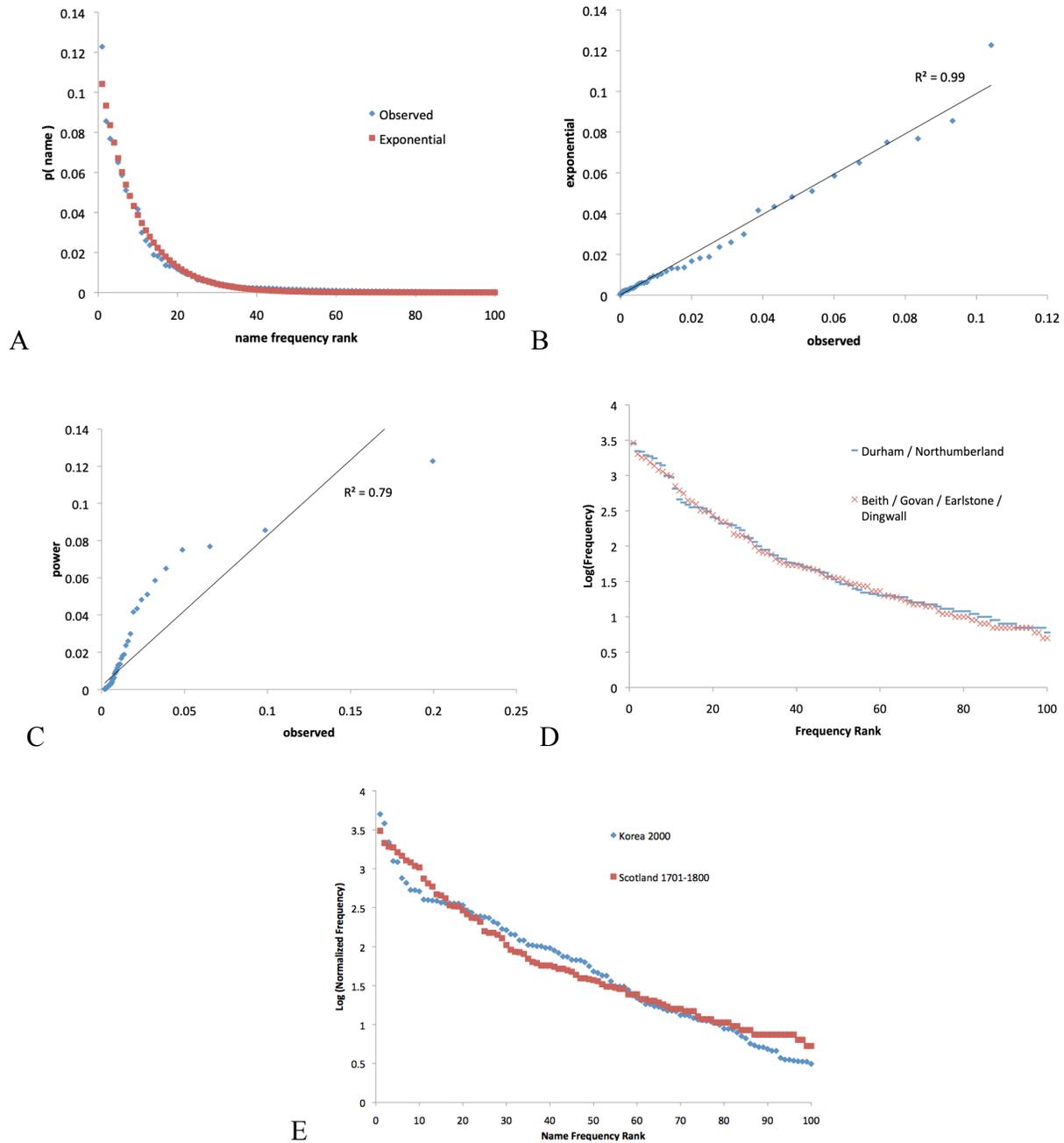

**Figure 3A**: 100 most frequent given names (98% of the population) 1701-1800 in 4 Scottish parishes plotted against an idealized exponential distribution. **B:** Pointwise comparison, observed (4.8 bits): idealized exponential (4.6 bits). **C:** Pointwise comparison, observed (4.8 bits): idealized power law distribution (5.3 bits). **D:** Log frequency x frequency rank plots of Durham and Northumberland name records 1701-1800 (4.8 bits) and the Scottish parishes (4.8 bits). **E:** Log (normalized) frequency x frequency rank plot of first names in South Korea 2000 (4.7 bits) and Scotland 1701-1800 (4.8 bits).

This raises a question: if English names still serve the same communicative function, why are they now Zipf distributed? The answer seems to lie in two facts: 1., mixtures of exponentials can produce power law distributions[9] [10] and 2., the US has a population of over 325 Million people, meaning that someone hearing a new name every 10 seconds for 12 hours a day would take over 200 years to hear every name. Since this greatly exceeds the human lifespan, it suggests that the set of all US first names is not an empirically relevant distribution. Rather, functional distributions are likely to be more localized, with names in Florida influencing the names given in Oregon less than other Oregonian names.

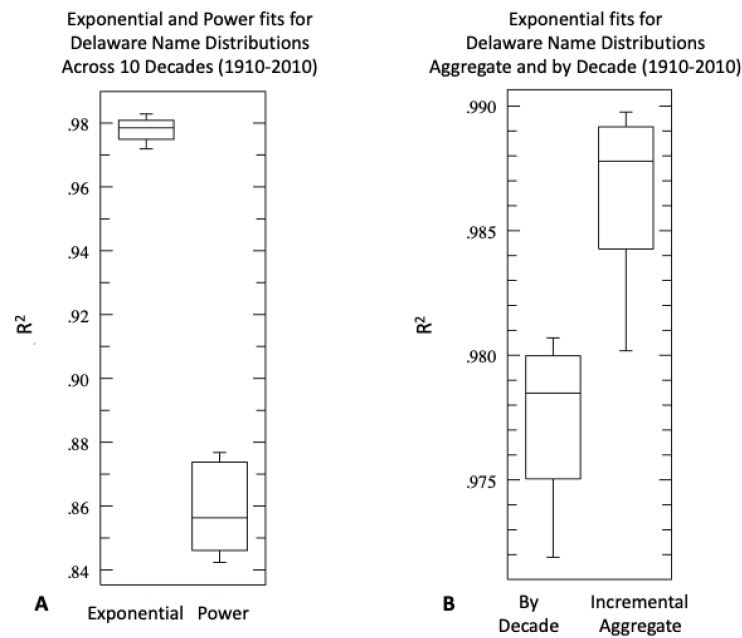

**Figure 4A**: Box plots of exponential and power law fits ($R^2$) for the Delaware first name distributions in each of 10 decades from 1910-2010. **B:** Exponential fit in each decade compared with the cumulative mixture at the end of each decade as the population is aggregated.

This hypothesis is supported by analyses of social security records in Delaware, one of the smallest States by area, with a small, densely distributed population. From 1910-2010 Delaware's by-decade first name distributions better fit an exponential than a power law, and incrementally mixing these samples (which are all taken from the same underlying distribution) improves this fit (Figure 4).

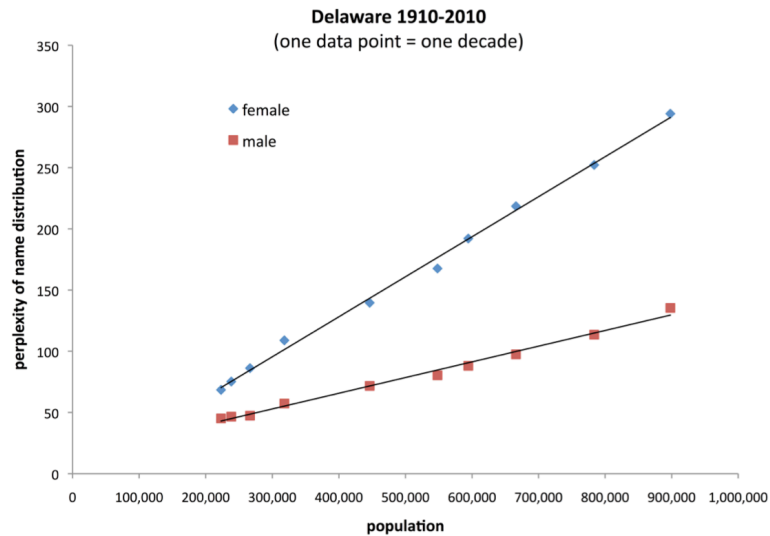

**Figure 5**: Increase in the information perplexity of the cumulative distribution of male and female names recorded in social security records in Delaware in each decade 1910-2010 plotted against population size (Both $R^2 > .99$).

Figure 5 plots the information perplexity of Delaware first names and population growth across the 20th Century, showing that the informativity of names given in each decade again appears to be a function of the information requirements of the name system. Figure 6 then shows that on average, the cumulative distributions of first names from 1910-2010 in each of the 50 US states and the District of Columbia also approximate exponentials, whereas the mixture of these different distributions better fits a power law. These findings support the suggestion that naming is sensitive to regional name information, and indicate that the power law distribution of the set of all US names is an artifact of mixing functionally relevant distributions [9] (further supporting the idea that power laws result from mixing functional distributions, smaller State population size correlated with a better exponential fit, $R^2 = .79$).

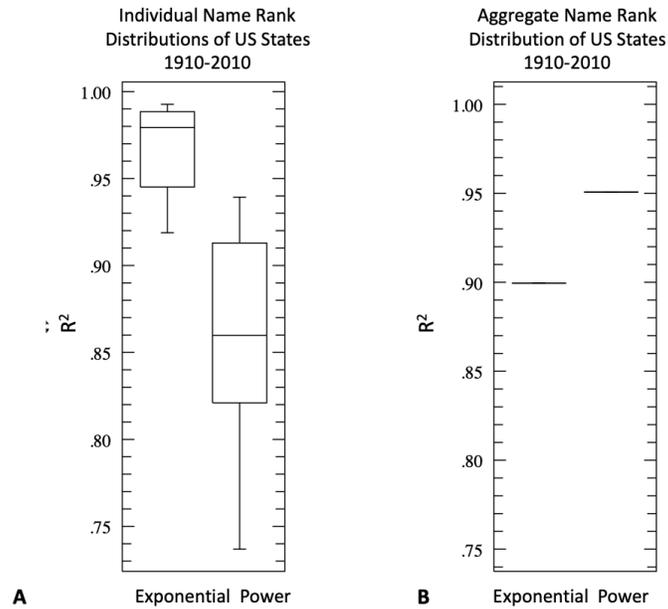

**Figure 6 A**: Box plots of exponential and power law fits ($R^2$) for the cumulative distributions of first names in each of the 50 US states and the District of Columbia, 1910-2010, **B:** Fits to a mixture of these distributions (i.e., the cumulative distribution of first names in the USA, 1910-2010).

Since these functional name distributions do not follow Zipf's law, this raises a question: Are name distributions unique, or do the Zipfian distribution observed in, say, English nouns and verbs[30] also result from the mixing of functional distributions? An obvious difference between nouns, verbs and first names is that distributions of the latter appear to be subcategorized geographically rather than grammatically (although any first name can occur after '*I'd like you to meet…*', the results presented above indicate that their probabilities are regionally conditioned), whereas the former are usually subcategorized by context. This occurs obviously in languages that have noun classes (i.e., in German, only nouns conditioned on the article *'der'* can be expected to occur after it), yet more subtly in the systematic alternation patterns that subcategorize verbs in English, where *"I am pleased to meet you…"* is grammatical, but *"I am liked to meet you…"* is not.

To explore the empirical properties of distributions of English verbs, the frequency distributions of 40 sets of published alternation patterns[31] were compared to those of verbs beginning with of the 20 most frequent English letters. As Figure 6 shows, whereas verbs following letters are Zipf distributed (like all verbs), the distributions of the subcategories of English verbs formed by alternation patterns appear to be geometric.

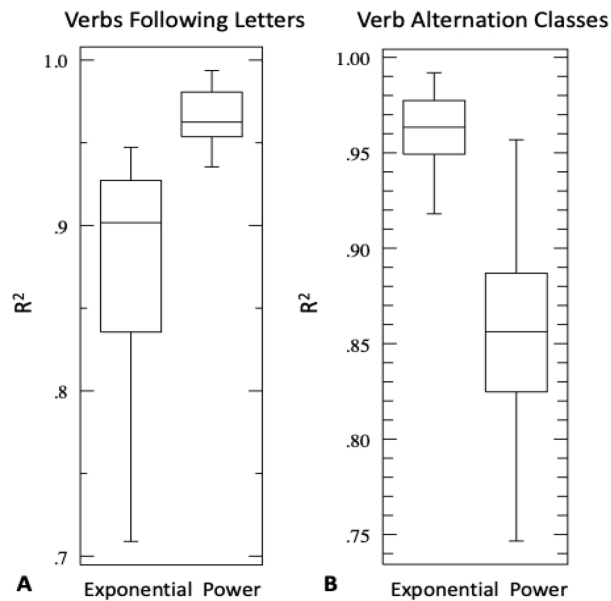

**Figure 6A**: Box plots of exponential and power law fits ($R^2$) for the distributions of 100 most frequent English verbs following the 20 most frequent English letters. **B:** Fits ($R^2$) for 40 English verb alternation classes.

Although English lacks noun classes, various techniques have been developed to use conditioning patterns in lexical distributions to yield coherent noun subcategories (i.e., sets of synonyms, or semantically related items). Given that the distributions discriminated by such patterns in child-focused speech are particularly relevant to questions about the learnability of languages, the generalizability of the findings described above to nouns was first assessed by an analysis of a test set of noun clusters shown to be learnable from a corpus of child / parent speech.[32] To compare these with adult speech, synonym sets for the test clusters were then produced using the same distributional methods.[33]

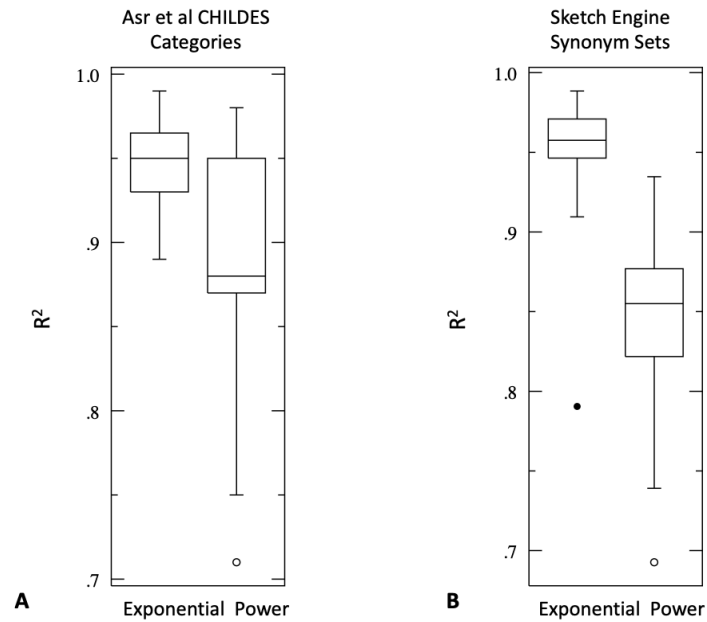

**Figure 7A**: Box plots of exponential and power law fits ($R^2$) for the distributions of 29 noun clusters extracted from a corpus of child directed speech. **B:** Fits ($R^2$) for 100 synonyms for the same classes.

Although the distribution of all English nouns follows Zipf's law,[33] these analyses confirmed that when broken down in to sub-categories based on the empirical contexts in which they occur, English nouns appear to be geometrically distributed, in both child-directed speech and adult generated text (Figure 7). When taken together with the analyses of name and verb distributions above, these results indicate that while empirical distributions of names are sub-categorized by physical geography, the topology of nouns and verbs is conceptually determined, with their empirical sub-categories being discriminated by the lexico-grammatical contexts in which they occur. Moreover, in every case it appears that the lexical items that comprise these empirical sub-categories are geometrically distributed.

Although numerous *post hoc* theories have sought to explain how Zipf's law serves to optimize human communication, information theory proves that the exponential family of distributions is better suited to this task. The analyses presented here confirm that the structure of the functional distributions of languages are consistent with the predictions of information theory rather than Zipf's law, which appears to be an artifact that arises when functional distributions are mixed.[10] [19] [34] While further study of these functional distributions is likely to advance the scientific understanding of human communicative capabilities in the future (and perhaps revolutionize our understanding of what human communication involves[35]), one of the results presented above exposes a problem at heart of current scientific practice. Consistent with many other communicative phenomena,[36] name grammars solve the problem of providing individuals with relatively unique identifiers in a remarkably efficient way.[21] [35] However, names are not always used in full form: scientific indexing, for example, is based on 'family names,' which are highly informative last names in the West, but relatively uninformative first names in the Sinosphere. This information imbalance massively increases the ambiguity of Sinosphere names in indexes, and has prompted at least one successful scientist to invert her name to offset it.[37] The data presented here strongly support this decision, because in showing how first names share a common information structure across languages, they reveal how and why it is that scientific indexes provide far less information about Sinosphere identities than Western identities. For Sinosphere authors, current practices are directly isomorphic to an index that renders the names Charles Darwin, Charles Dodgson and Charles Dickens all as Charles, D. The quantification of name information presented here not only highlights this problem, it may even offer a formal basis for its solution.